# Robust Multilingual Text-to-Pictogram Mapping for Scalable Reading Rehabilitation

Soufiane Jhilal
*Sony Computer Sciences Laboratories*
Paris, France
soufiane.jhilal@sony.com

Martina Galletti
*Sony Computer Sciences Laboratories*
Paris, France
martina.galletti@sony.com

***Abstract*—Reading comprehension presents a significant challenge for children with Special Educational Needs and Disabilities (SEND), often requiring intensive one-on-one reading support. To assist therapists in scaling this support, we developed a multilingual, AI-powered interface that automatically enhances text with visual scaffolding. This system dynamically identifies key concepts and maps them to contextually relevant pictograms, supporting learners across languages. We evaluated the system across five typologically diverse languages (English, French, Italian, Spanish, and Arabic), through multilingual coverage analysis, expert clinical review by speech therapists and special education professionals, and latency assessment. Evaluation results indicate high pictogram coverage and visual scaffolding density across the five languages. Expert audits suggested that automatically selected pictograms were semantically appropriate, with combined correct and acceptable ratings exceeding 95% for the four European languages and approximately 90% for Arabic despite reduced pictogram repository coverage. System latency remained within interactive thresholds suitable for real-time educational use. These findings support the technical viability, semantic safety, and acceptability of automated multimodal scaffolding to improve accessibility for neurodiverse learners.**

## I. INTRODUCTION

Inclusive education increasingly depends on learning technologies that can adapt to diverse learner profiles, especially for children with Special Educational Needs and Disabilities (SEND). This is particularly critical in reading, as text remains the dominant medium for instruction and assessment across educational contexts. Reading is a complex cognitive process requiring the integration of textual information, prior knowledge, and contextual cues to construct meaning [1], a challenge for many neurodiverse children who often experience deficits in auditory/visual processing, language skills, attention, and working memory [2]. Traditional interventions for these learners typically involve one-on-one sessions with specialized therapists, which, while effective, are resource-intensive and difficult to scale due to limited professional availability [3]. Consequently, many SEND children face delayed or infrequent support, widening educational gaps and underscoring the urgent need for scalable digital solutions to ensure equitable access to reading interventions.

AI-powered interventions offer a promising alternative by delivering consistent, personalized support that complements traditional therapy while overcoming geographic and resource constraints [4]. To this end, we have developed ARTIS [5], [6], an AI-powered digital interface specifically designed to support text comprehension rehabilitation. ARTIS dynamically extracts pictograms from keywords, identifies difficult words, generates concept maps, and provides targeted exercises. This multimodal approach offers tailored support and helps close critical gaps in educational provision. This paper presents a multilingual extension of the interface, and validates its technical viability and semantic alignment through multilingual coverage analysis, clinician audit, and latency evaluation.

## II. RELATED WORK

Scalable and accessible tools for neurodiverse learners are in great need. While AI-powered innovations have transformed possibilities across education [7] and clinical settings [8], [9], relatively few efforts have focused on reading and language comprehension technologies explicitly designed for neurodiverse populations, and existing approaches remain limited in scope, personalization, or scalability.

### *A. AI-powered Reading Comprehension Tools*

Digital platforms increasingly use AI to support reading through summarization, question-answering, simplification, and semantic annotation. Tools such as Semantic Reader [10] and SciReader [11] serve general users with semantic annotations and contextual summaries but lack adaptations for individuals with comprehension deficits.

Other approaches, such as gaze-driven interfaces [12], adapt content difficulty using eye-tracking and NLP, while educational platforms like 3D Readers [13] and CACSR [14] apply instructional strategies. However, none integrate the multimodal, therapeutic support needed for SEND learners. The RIDInet [15] platform represents a specialized effort in Italian telerehabilitation, providing inferential comprehension exercises but lacking keyword or pictogram integration.

### *B. Keyword Extraction in Educational Contexts*

A key component of effective reading support is identifying important concepts in text. Keyword extraction (KE) focuses on isolating relevant terms that capture the core meaning. It has shown benefits for reading comprehension and vocabulary learning [16], [17]; however, its use in adaptive, multimodal learning environments remains limited. Few studies explore personalized keyword generation during reading or compare KE with other comprehension strategies [18]. For children with SEND, who often struggle to identify key information due to cognitive or linguistic challenges, automated KE can focus attention and reduce cognitive load.

### *C. Pictogram Integration and Multimodal Learning*

While KE highlights important concepts, SEND learners often require additional visual scaffolding to bridge the gap between textual information and comprehension. Students with dyslexia and language disorders rely on non-verbal and visual-spatial abilities to compensate for linguistic difficulties [19], [20], [21], making multimodal interventions valuable. Simplified visual symbols like pictograms can facilitate comprehension in individuals with language disorders by providing concrete, easily interpretable representations that reduce reliance on linguistic processing [22].

Recent pictogram systems have used rule-based translation with NLP enhancements like word-sense disambiguation [23]. French extensions [24] and deep-learning approaches [25] have improved coverage and contextual accuracy; however, educational applications remain limited, particularly for rehabilitation. Unlike general multimodal scaffolding approaches, our system specifically addresses SEND learners' needs through sentence segmentation with contextually disambiguated pictogram augmentation.

## III. System Overview

ARTIS is a reading support platform designed for supervised educational or rehabilitation sessions, where an adult facilitator (teacher, therapist, or trained operator) can guide the learner and adjust assistance to session goals. The platform can present text with adjustable supports including sentence-level presentation (structural support), keyword emphasis (lexical support), pictograms aligned to selected terms (semantic support), and optional audio playback of sentences. In addition to in-text scaffolding, ARTIS provides on-demand word definitions and targeted vocabulary and grammatical exercises derived from the reading material, enabling active reinforcement of linguistic concepts. This adjustable design reflects a practical reality in SEND contexts: the right scaffold level is not constant across learners or even within a session. Figure 1 below shows an example from the interface where a sentence is presented with highlighted keywords and pictogram augmentation.

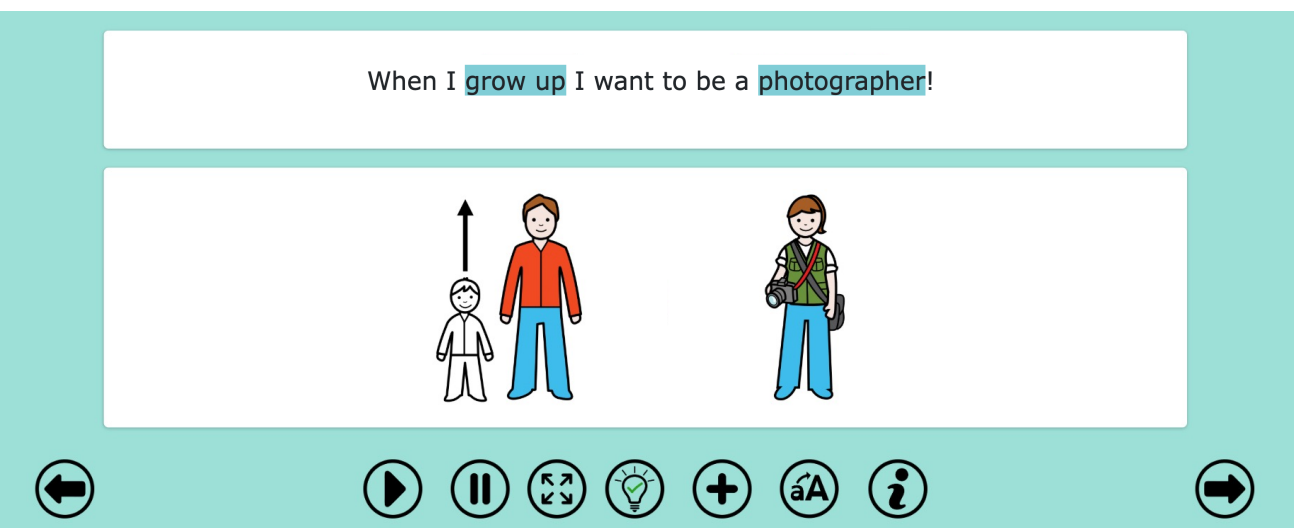

**Figure 1**: Screenshot from the interface showing a single sentence presented during sequential reading, with highlighted keywords and pictogram support.

### A. Pictogram Repository Selection

We initially explored using generative AI (e.g., diffusion models fine-tuned on pictograms) to create custom visual icons. However, early tests showed drawbacks such as inconsistent style, unclear semantics, and risk of cognitive overload. We therefore prioritized a repository-based approach that relies on pictograms from a well-established and clinically validated source, ensuring both consistency and clinical reliability.

We adopted ARASAAC [26], a free and multilingual pictogram repository widely used in special education and Augmentative and Alternative Communication (AAC) contexts, as our pictogram source. ARASAAC offers a large set of AAC symbols used by families and professionals. Its terms of use state that resources and derived materials are published under CC BY-NC-SA, permitting reuse for non-profit purposes with attribution and share-alike requirements.

While other repositories exist, some are field-specific, such as SantéBD [27] for the medical field, while others are behind a paywall, like BeTa [28]. Mulberry [29] and Sclera [30] are two of the largest and most widely used repositories; however, they don't offer the same coverage as ARASAAC. Mulberry provides 3,436 pictograms covering 3,119 keywords in English, with some initiatives to map them to other languages like French and Finnish. Sclera offers 12,892 pictograms, mainly in black and white, covering 3,119 keywords in English, with relatively similar coverage in Spanish, French, German, and Dutch, and substantially less coverage (around half) in Polish. In contrast, ARASAAC offers a more extensive repository of 13,709 pictograms in 38 languages, covering over 18,000 keywords in some of them (see Figure 2). Even more importantly, ARASAAC is the only repository that provides extensive metadata with the definition of each pictogram to avoid polysemy cases.

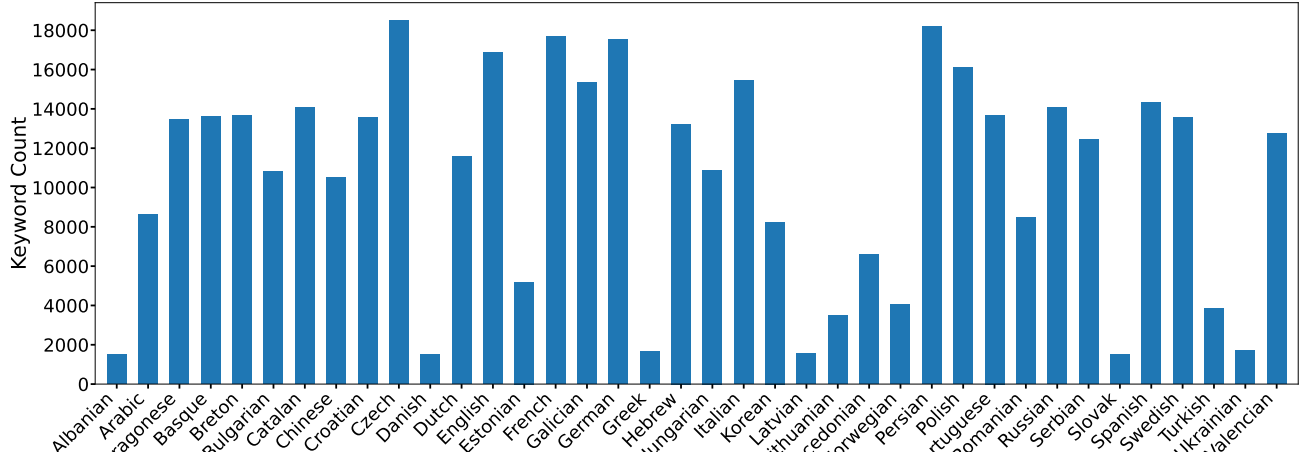

**Figure 2:** Number of keywords with associated pictograms available in ARASAAC across 38 languages.

### B. Text-to-Pictogram Pipeline

While ARTIS includes multiple interface features, one of the central enabling components is the text-to-pictogram pipeline that identifies candidate concepts in text and maps them to relevant pictograms. We developed a multilingual, and context-aware pipeline (Figure 3) that integrates language detection, keyword extraction, semantic disambiguation, and pictogram retrieval. Below we detail each step of the pipeline:

1. ***Language Detection and Sentence Segmentation***: The pipeline detects the input language using the Lingua [31] detector and segments text into sentences using NLTK's tokenizer, tailored for the detected language. This ensures language-agnostic compatibility beyond the Italian-focused configuration in prior work.

2. ***Tokenization and Lemmatization***: Sentences undergo tokenization and lemmatization via a cascading strategy:

• Primary tokenization and lemmatization are performed by language-specific spaCy models, which handle multiword expressions effectively.

• If spaCy models are unavailable for a detected language, the pipeline defaults to simplemma for fallback coverage, ensuring robustness across diverse linguistic contexts. This cascading approach enhances reliability and supports broad multilingual coverage.

3. ***Comprehensive Keyword Extraction***: Keywords for pictogram selection are extracted using YAKE [32]. It outputs a ranked list of keywords (including multiword expressions: keyphrases); we retain the top three candidates per sentence.

4. ***Contextual Retrieval and Semantic Matching of Pictograms***: Each keyword is queried against ARASAAC, either via the public API or via a locally cached snapshot of ARASAAC metadata. The system retrieves multiple candidate pictograms, each accompanied by language-specific semantic definitions. Since keywords can have multiple meanings, we add a new semantic disambiguation module using transformer-based embeddings (via the SentenceTransformers library [33]):

• A local sentence context (±2-word window around the keyword) is embedded using language-appropriate sentence transformers (e.g., multilingual MiniLM).

• Embeddings of each pictogram definition provided by ARASAAC are computed either on-the-fly, or precomputed offline when using the cached snapshot.

• The pictogram definition with the highest cosine similarity to the local context embedding is selected, supporting automated disambiguation and reducing reliance on manual selection.

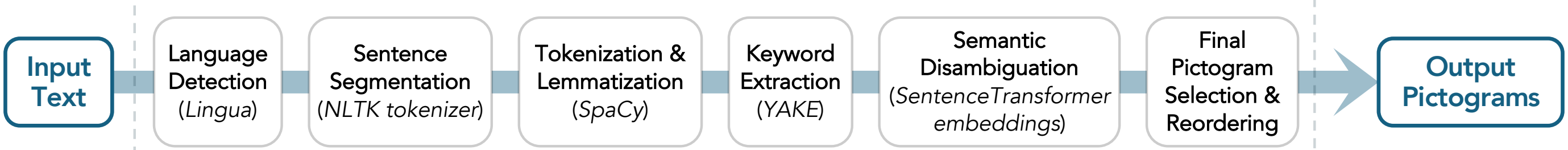

**Figure 3**: Overview of the text-to-pictogram pipeline.

• For numeric keywords, a simpler heuristic is employed, automatically selecting a straightforward numeric pictogram to ensure accurate representation.

5. ***Ordered Output and Integration***: Selected pictograms are reordered to match their original textual sequence, enhancing visual coherence and comprehension. Results are dynamically integrated into the ARTIS interface, aligning pictograms with corresponding textual keywords.

## IV. System Evaluation

To evaluate our system, we conducted a multi-layered evaluation focusing on multilingual scalability, pictogram coverage, semantic accuracy, and system latency.

We selected *Le Petit Prince* (The Little Prince) [34] as our primary evaluation corpus. While ARTIS has been previously tested on multiple texts and studies, especially in Italian, this study aims to validate the new multilingual pipeline on multiple languages. *Le Petit Prince* is widely available across many languages, enabling reproducible multilingual evaluation. Moreover, its narrative structure and vocabulary align with the reading level of our primary target demographic (primary school children). We used legally obtained digital versions of *Le Petit Prince* for offline analysis only and do not redistribute the corpus; we report only aggregate statistics.

We evaluated the pipeline across five languages: English (EN), French (FR), Italian (IT), Spanish (ES), and Arabic (AR). This selection provides linguistic diversity (Romance languages, Germanic, and Semitic), script variation (Latin and Arabic scripts testing pipeline robustness), substantial ARASAAC repository coverage (see Figure 2), and global relevance as these languages collectively represent around 2.8 billion speakers [35] in regions where ARTIS is likely to be deployed. Critically, this selection was also constrained by the availability of qualified clinical reviewers (speech therapists and special education professionals) who could provide expert acceptability judgments in each language.

### A. Pictogram Coverage

First, we assessed the system's ability to provide pictogram support using two metrics capturing keyword success and sentence-level utility.

**Keyword to pictogram (K2P) coverage** measures the percentage of extracted keywords that successfully retrieve at least one pictogram candidate.

**Sentence level coverage** evaluates the distribution of sentences by pictogram density: (1) zero pictograms, (2) partial coverage (some keywords mapped), and (3) full coverage (all keywords mapped). We also report mean pictograms per sentence to quantify visual scaffolding density, capturing the practical utility beyond individual keyword success rates.

### B. Clinical Acceptability

Given our vulnerable target population, we conducted a clinician-led audit evaluating whether the system's automatic pictogram selection is semantically safe and pedagogically appropriate. *Le Petit Prince* contains 1,855 sentences in the English version, with comparable length across the other languages. We randomly picked a subsample of 200 sentences for this evaluation, where native-speaking speech therapists and special education professionals (2 reviewers for English, 5 for French, 2 for Italian, 3 for Spanish, and 2 for Arabic) reviewed each sentence presented with the highlighted keywords and the selected pictograms. Reviewers rated each keyword and pictogram according to Table I below:

**TABLE I:** RATING SCALE FOR KEYWORD AND PICTOGRAM EVALUATION

| Rating | Keywords | Pictograms |
|---|---|---|
| Correct (C) | Key concept that aids understanding | Accurately represents intended meaning |
| Acceptable (A) | Somewhat helpful but not essential | Reasonable though not optimal |
| Incorrect (I) | Unhelpful or irrelevant for comprehension support | Mismatched or misleading |

### C. System Deployability

To validate the system's suitability for real-world deployment we evaluated end-to-end latency under realistic settings: ARASAAC candidate lists and definitions are served from a local cache, and pictogram-definition embeddings are precomputed offline. At runtime, the system computes the sentence-context embedding for each keyword and performs cosine-similarity matching. We measure processing time per sentence on a standard CPU laptop.

## V. Results

### A. Pictogram Coverage

Table II presents coverage statistics across all five languages. The system extracted an average of 2.55–2.70 keywords per sentence across languages, demonstrating consistent keyword identification regardless of linguistic structure.

**TABLE II:** COVERAGE METRICS ACROSS LANGUAGES

| Metric | EN | FR | IT | ES | AR |
|---|---|---|---|---|---|
| Mean keywords/sentence | 2.62 | 2.70 | 2.65 | 2.60 | 2.55 |
| K2P coverage (%) | 97.9 | 94.2 | 93.8 | 93.9 | 79.4 |
| Mean pictograms/sentence | 2.56 | 2.54 | 2.49 | 2.44 | 2.02 |
| Sentences: 0 pictograms (%) | 2.8 | 4.4 | 3.6 | 4.1 | 9.7 |
| Sentences: partial (%) | 20.6 | 27.4 | 24.2 | 26.1 | 48.8 |
| Sentences: full (%) | 76.6 | 68.2 | 72.2 | 69.8 | 41.5 |

Keyword to pictogram coverage ranged from 79.4% (Arabic) to 97.9% (English), with Romance and Germanic languages achieving over 93% coverage. This variation reflects differences in ARASAAC repository depth across languages (see Figure 2). At the sentence level, the system provided substantial visual scaffolding, averaging 2.02–2.56 pictograms. Sentences with full pictogram coverage accounted for 41.5–76.6% of all sentences, with higher rates observed in European languages (EN, FR, IT, ES) compared to Arabic. Only 2.8–9.7% of sentences received zero pictogram support, indicating broad utility across the corpus. The majority of remaining sentences received partial coverage, where at least some keywords were visually supported.

### B. Clinical Acceptability

The clinician audit (Table III) indicates high semantic alignment. Keyword extraction achieved high relevance, with "Correct" ratings ranging from 72.5% (FR) to 78.0% (EN) across European languages. When including "Acceptable" ratings, keyword utility exceeded 90% for these languages. Arabic showed slightly lower performance (66.0% correct keywords) due to linguistic complexity.

**TABLE III:** CLINICIAN RATINGS OF EXTRACTED KEYWORDS AND SELECTED PICTOGRAMS ACROSS LANGUAGES. PERCENTAGES INDICATE PROPORTIONS OF CORRECT (C), ACCEPTABLE (A), AND INCORRECT (I) JUDGMENTS.

| Reviewer | Keywords rated | Pictograms rated | Keywords | | | Pictograms | | |
|---|---|---|---|---|---|---|---|---|
| | | | C (%) | A (%) | I (%) | C (%) | A (%) | I (%) |
| EN1 | 524 | 506 | 79.2 | 16.0 | 4.8 | 87.0 | 10.4 | 2.6 |
| EN2 | 524 | 506 | 76.8 | 17.6 | 5.6 | 86.0 | 10.8 | 3.2 |
| FR1 | 541 | 509 | 73.8 | 18.5 | 7.7 | 83.5 | 12.0 | 4.5 |
| FR2 | 541 | 509 | 71.2 | 20.0 | 8.8 | 80.8 | 13.7 | 5.5 |
| FR3 | 541 | 509 | 73.0 | 19.0 | 8.0 | 82.6 | 12.9 | 4.5 |
| FR4 | 541 | 509 | 70.5 | 20.7 | 8.8 | 81.1 | 13.9 | 5.0 |
| FR5 | 541 | 509 | 73.9 | 19.3 | 6.8 | 82.0 | 13.0 | 5.0 |
| IT1 | 532 | 497 | 76.3 | 17.5 | 6.2 | 85.2 | 11.8 | 3.0 |
| IT2 | 532 | 497 | 74.5 | 18.5 | 7.0 | 83.4 | 13.0 | 3.6 |
| ES1 | 527 | 488 | 75.1 | 18.0 | 6.9 | 84.2 | 12.1 | 3.7 |
| ES2 | 527 | 488 | 73.0 | 19.2 | 7.8 | 82.3 | 13.0 | 4.7 |
| ES3 | 527 | 488 | 73.9 | 18.9 | 7.2 | 82.8 | 13.3 | 3.9 |
| AR1 | 519 | 435 | 66.8 | 19.4 | 13.8 | 75.0 | 15.2 | 9.8 |
| AR2 | 519 | 435 | 65.2 | 20.6 | 14.2 | 73.4 | 16.0 | 10.6 |
| **Averages per language** | | | | | | | | |
| EN | 524 | 506 | 78.0 | 16.8 | 5.2 | 86.5 | 10.6 | 2.9 |
| FR | 541 | 509 | 72.5 | 19.5 | 8.0 | 82.0 | 13.1 | 4.9 |
| IT | 532 | 497 | 75.4 | 18.0 | 6.6 | 84.3 | 12.4 | 3.3 |
| ES | 527 | 488 | 74.0 | 18.7 | 7.3 | 83.1 | 12.8 | 4.1 |
| AR | 519 | 435 | 66.0 | 20.0 | 14.0 | 74.2 | 15.6 | 10.2 |

Pictogram selections received higher acceptability ratings than keyword relevance ratings, with the combined "Correct" and "Acceptable" rate exceeding 95% across European languages. This indicates that even when a keyword was deemed only somewhat helpful (Acceptable), the associated pictogram often remained semantically appropriate. English achieved the highest accuracy (86.5% Correct), while Arabic showed slightly lower performance (74.2% Correct, and ~90% combined utility) potentially due to morphological complexity. Crucially for clinical safety, the "Incorrect" rate for pictograms remained low, ranging from 2.9% (English) to 4.9% (French) in European languages, and 10.2% for Arabic due to reduced repository coverage and morphological complexity. This shows that the system rarely introduces misleading visual stimuli, minimizing the risk of misconception during semi-autonomous practice.

### *C. System Deployability*

Table IV and Figure 4 summarize the system's end-to-end processing latency across languages.

**TABLE IV:** SUMMARY STATISTICS OF END-TO-END SENTENCE PROCESSING LATENCY (IN MILLISECONDS) ACROSS LANGUAGES.

| Language | Mean | Median | SD | 95th perc. | Min | Max |
|---|---|---|---|---|---|---|
| EN | 176.26 | 160.35 | 57.02 | 303.10 | 99.01 | 556.52 |
| FR | 192.27 | 173.35 | 61.09 | 329.00 | 127.02 | 593.81 |
| IT | 178.01 | 160.45 | 57.25 | 309.18 | 123.96 | 564.93 |
| ES | 185.94 | 167.83 | 59.26 | 317.35 | 124.98 | 563.05 |
| AR | 228.07 | 206.74 | 71.50 | 396.53 | 148.66 | 675.50 |

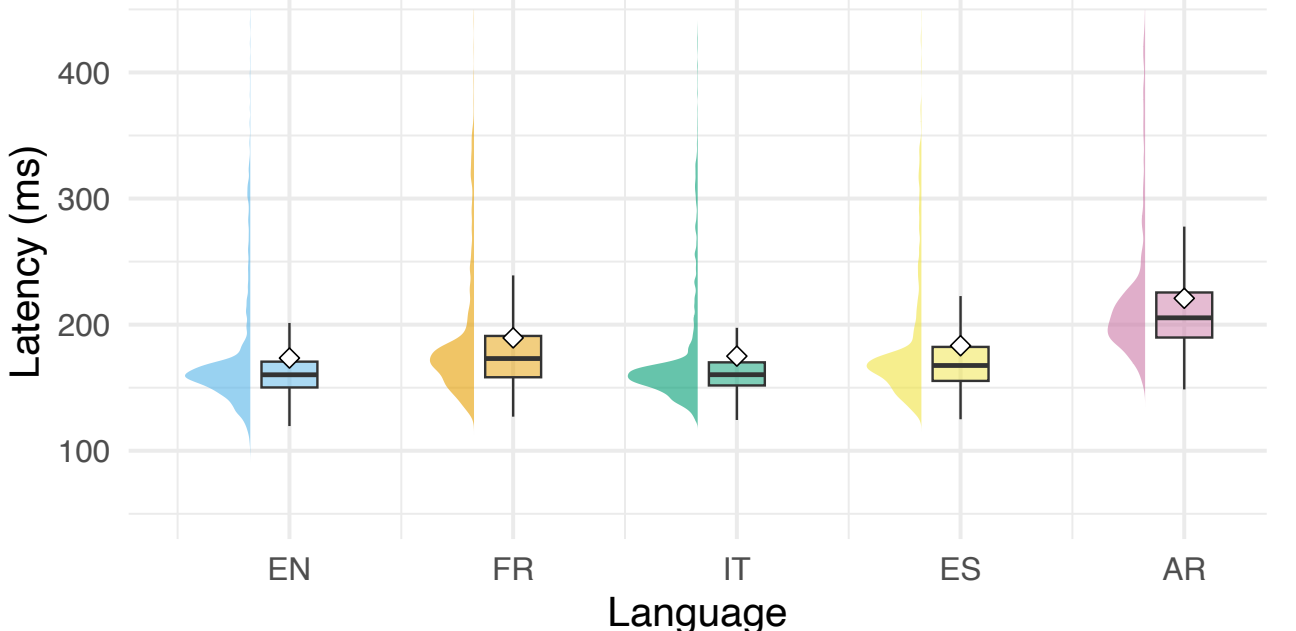

**Figure 4**: Distribution of end-to-end sentence processing latency (in milliseconds) across languages.

Mean processing times ranged from 176 ms (English) to 228 ms (Arabic). For most languages, latency remained below or near the 200 ms threshold for perceived instantaneity (see Figure 4 for distributions). Even with 95th percentile peaks around 300–400 ms, the visual scaffolding appears without disrupting the reading flow. The slightly higher latency in Arabic reflects the computational cost of processing its complex morphology but remains well within acceptable limits for educational applications.

## VI. DISCUSSION

This study validated the technical viability and semantic alignment of a multilingual text-to-pictogram pipeline embedded in ARTIS. Across five languages, the system extracted a similar number of keywords per sentence (2.55–2.70), suggesting that the keyword extraction strategy generalizes across typologically different languages. High keyword to pictogram coverage in English, French, Italian, and Spanish (>93%) indicates that for languages with strong ARASAAC metadata support, pictogram augmentation can be provided at scale with minimal uncovered concepts. Sentence-level results further show dense scaffolding (2.44–2.56 pictograms/sentence) with only a small proportion of sentences receiving no pictograms (2.8–4.4%), supporting practical use in sequential reading workflows.

Arabic showed lower keyword to pictogram coverage (79.4%), higher partial sentence coverage (48.8%), and a higher incorrect pictogram rate (10.2%). This likely reflects differences in ARASAAC coverage across languages and Arabic's morphological complexity, which can reduce exact matches and increase ambiguity. This gap also highlights a broader deployment constraint, as multilingual assistive scaffolds can be limited by language-resource depth and may disproportionately affect learners in under-resourced language contexts. Despite these challenges, the combined "Correct" and "Acceptable" pictogram rate in Arabic remained close to 90%, suggesting that the pipeline can still provide useful visual support, albeit with a higher need for facilitator oversight.

Clinician ratings provide an initial content validity signal as incorrect pictograms remained below 5% in European languages and around 10% in Arabic, indicating that the disambiguation approach typically avoids misleading representations. Most incorrect mappings were due to polysemy-related sense disambiguation errors and abstract concepts lacking suitable pictograms, with very few errors arising from named entities or preprocessing issues. Keyword relevance ratings were lower than pictogram ratings, which is expected as keyword extraction optimizes salience rather than instructional necessity. This motivates future refinements that incorporate disability-sensitive selection criteria and more detailed error analyses.

Latency results indicate interactive performance with cached ARASAAC metadata and offline precomputed embeddings (176–228 ms mean; 95th percentile 303–397 ms). Higher Arabic latency likely reflects additional processing costs associated with handling a morphologically complex script and increased ambiguity during text processing but remained within acceptable bounds for educational use. The cached configuration also supports reliable operation where connectivity is inconsistent, which is common in school and clinical environments.

While these results establish technical feasibility and semantic appropriateness across diverse linguistic contexts, some limitations warrant acknowledgment. Clinician ratings, though valuable for initial validation, cannot substitute for direct evaluation with SEND learners in authentic educational settings. Initial evidence from a companion user study suggests that responses to multimodal scaffolding are heterogeneous: while many SEND learners benefit from the full combination of supports, others engage more effectively with select modalities such as using only highlighted keywords or only pictograms to optimize their attention [36]. This aligns with ARTIS's core design philosophy of integrating a human-in-the-loop capability that allows therapists to adjust or toggle scaffolding levels rather than relying on a static, fully automated approach. Another limitation is the system's dependence on ARASAAC repository coverage creates inherent equity gaps across languages, underscoring the need for community-driven expansion of pictogram resources. Addressing these limitations will be essential to realizing the broader goal of truly inclusive, technology-supported reading interventions that equitably serve neurodiverse learners regardless of linguistic background or cognitive profile.

## VII. Conclusion

This work presented and validated a multilingual extension of ARTIS, an AI-powered reading support platform for children with SEND. By demonstrating that automated multimodal scaffolding can scale across linguistically diverse contexts while maintaining clinical appropriateness, this study addresses core challenges in accessible, technology-supported education.

The system's reliance on open, validated AAC resources and its language-agnostic architecture exemplify how AI can remove traditional learning barriers and enable equitable participation for neurodiverse learners. However, reduced performance in Arabic underscores a key challenge: extending assistive technologies to under-resourced languages risks perpetuating educational inequities precisely where inclusive tools are most needed. This highlights the urgent need for continued investment in multilingual accessibility to ensure truly inclusive digital learning ecosystems that leave no learner behind.

Future work should include longitudinal studies with target users in authentic educational settings to assess pedagogical effectiveness and learning outcomes as natural language processing capabilities and multilingual AAC repositories continue to expand.

## References

[1] C. Hulme and M. J. Snowling, "Learning to Read: What We Know and What We Need to Understand Better," *Child Dev Perspect*, vol. 7, no. 1, pp. 1–5, Mar. 2013, doi: 10.1111/cdep.12005.

[2] H. Scarborough, F. Fletcher-Campbell, J. Soler, and G. Reid, "Connecting early language and literacy to later reading (dis)abilities: Evidence, theory, and practice," *Approaching difficulties in literacy development: assessment, pedagogy, and programmes*, pp. 23–39, Jan. 2009.

[3] K. Squires, "Addressing the Shortage of Speech-Language Pathologists in School Settings," *Journal of the American Academy of Special Education Professionals*, pp. 131–137, Feb. 2013.

[4] Y. Yang, L. Chen, W. He, D. Sun, and S. Z. Salas-Pilco, "Artificial Intelligence for Enhancing Special Education for K-12: A Decade of Trends, Themes, and Global Insights (2013–2023)," *Int J Artif Intell Educ*, vol. 35, no. 3, pp. 1129–1177, Sep. 2025, doi: 10.1007/s40593-024-00422-0.

[5] M. Galletti *et al.*, "A Reading Comprehension Interface for Students with Learning Disorders," in *International Conference on Multimodal Interaction*, Paris France: ACM, Oct. 2023, pp. 282–287. doi: 10.1145/3610661.3616176.

[6] M. Galletti, E. Pasqua, M. Calanca, C. Marchesi, D. Tomaiuoli, and D. Nardi, "ARTIS: a digital interface to promote the rehabiliatation of text comprehension difficulties through Artificial Intelligence," presented at the Ital-IA 2024: 4th National Conference on Artificial Intelligence, 2024.

[7] S. Wang, F. Wang, Z. Zhu, J. Wang, T. Tran, and Z. Du, "Artificial intelligence in education: A systematic literature review," *Expert Systems with Applications*, vol. 252, p. 124167, Oct. 2024.

[8] Y. Zhang, Y. Weng, and J. Lund, "Applications of Explainable Artificial Intelligence in Diagnosis and Surgery," *Diagnostics*, vol. 12, no. 2, p. 237, Feb. 2022, doi: 10.3390/diagnostics12020237.

[9] S. Jhilal, S. Marchesotti, B. Thirion, B. Soudrie, A.-L. Giraud, and E. Mandonnet, "Implantable Neural Speech Decoders: Recent Advances, Future Challenges," *Neurorehabil Neural Repair*, Sep. 2025, doi: 10.1177/15459683251369468.

[10] K. Lo *et al.*, "The Semantic Reader Project," *Commun. ACM*, vol. 67, no. 10, pp. 50–61, Sep. 2024, doi: 10.1145/3659096.

[11] A. Head *et al.*, "Augmenting Scientific Papers with Just-in-Time, Position-Sensitive Definitions of Terms and Symbols," in *Proceedings of the 2021 CHI Conference on Human Factors in Computing Systems*, in CHI '21. New York, NY, USA: Association for Computing Machinery, May 2021, pp. 1–18.

[12] T. Higasa, K. Tanaka, Q. Feng, and S. Morishima, "Gaze-Driven Sentence Simplification for Language Learners: Enhancing Comprehension and Readability," in *International Cconference on Multimodal Interaction*, Oct. 2023, pp. 292–296. doi: 10.1145/3610661.3616177.

[13] M. C. Johnson-Glenberg, "Web-based reading comprehension instruction: Three studies of 3D-readers," in *Reading comprehension strategies: Theories, interventions, and technologies*, Mahwah, NJ, US: Lawrence Erlbaum Associates Publishers, 2007, pp. 293–324.

[14] A.-H. Kim, S. Vaughn, J. K. Klingner, A. L. Woodruff, C. Klein Reutebuch, and K. Kouzekanani, "Improving the Reading Comprehension of Middle School Students With Disabilities Through Computer-Assisted Collaborative Strategic Reading," *Remedial and Special Education*, vol. 27, no. 4, pp. 235–249, Jul. 2006.

[15] "RIDInet." Available: https://www.anastasis.it/ridinet/

[16] M. A. McDaniel and M. Pressley, "Keyword and context instruction of new vocabulary meanings: Effects on text comprehension and memory," *Journal of Educational Psychology*, vol. 81, no. 2, pp. 204–213, 1989, doi: 10.1037/0022-0663.81.2.204.

[17] Y. Seki, K. Akahori, and T. Sakamoto, "Using Key Words to Facilitate Text Comprehension," *Educational technology research*, vol. 16, no. 1–2, pp. 11–21, 1993.

[18] D. M. Chun and J. L. Plass, "Research on text comprehension in multimedia environments," Jul. 1997, doi: 10.64152/10125/25004.

[19] E. Danis, A.-M. Nader, J. Degré-Pelletier, and I. Soulières, "Semantic and Visuospatial Fluid Reasoning in School-Aged Autistic Children," *J Autism Dev Disord*, vol. 53, no. 12, pp. 4719–4730, Dec. 2023, doi: 10.1007/s10803-022-05746-1.

[20] S. Jhilal, N. Molinaro, and A. Klimovich-Gray, "Non-verbal skills in auditory word processing: implications for typical and dyslexic readers," *Language, Cognition and Neuroscience*, vol. 40, no. 3, pp. 341–359, Mar. 2025, doi: 10.1080/23273798.2024.2438012.

[21] L. Superbia-Guimarães, M. Bader, and V. Camos, "Can children and adolescents with ADHD use attention to maintain verbal information in working memory?," *PLOS ONE*, vol. 18, no. 3, p. e0282896, Mar. 2023, doi: 10.1371/journal.pone.0282896.

[22] C. Vaschalde, P. Trial, E. Esperança-Rodier, D. Schwab, and B. Lecouteux, "Automatic pictogram generation from speech to help the implementation of a mediated communication," in *Conference on Barrier-free Communication*, Geneva, Switzerland, Nov. 2018.

[23] L. Sevens, G. Jacobs, V. Vandeghinste, I. Schuurman, and F. Van Eynde, "Improving Text-to-Pictograph Translation Through Word Sense Disambiguation," in *Proceedings of the Fifth Joint Conference on Lexical and Computational Semantics*, C. Gardent, R. Bernardi, and I. Titov, Eds., Berlin, Germany: Association for Computational Linguistics, Aug. 2016, pp. 131-135.

[24] M. Norré, V. Vandeghinste, P. Bouillon, and T. François, "Extending a Text-to-Pictograph System to French and to Arasaac," in *Proceedings of the International Conference on Recent Advances in Natural Language Processing (RANLP 2021)*, Sep. 2021, pp. 1050–1059.

[25] J. A. Pereira, D. Macêdo, C. Zanchettin, A. L. I. de Oliveira, and R. do N. Fidalgo, "PictoBERT: Transformers for next pictogram prediction," *Expert Systems with Applications*, vol. 202, p. 117231, Sep. 2022.

[26] "ARASAAC." Available: https://beta.arasaac.org/

[27] "SantéBD." Available: https://santebd.org/

[28] "Beta Symbols." Available: https://www.betasymbols.com/en/

[29] "Mulberry Symbols." Available: https://mulberrysymbols.org/

[30] "Sclera symbols." Available: https://www.sclera.be/en/vzw/home

[31] "Lingua." Available: https://github.com/pemistahl/lingua-py

[32] "YAKE." Available: https://github.com/INESCTEC/yake

[33] "SentenceTransformers." Available: https://sbert.net/

[34] A. de Saint-Exupéry, *Le Petit Prince*. Gallimard, 1943.

[35] C. D. Fennig, *Ethnologue: Languages of the Americas and the Pacific: 28*. S.l.: Sil International, Global Publishing, 2025.

[36] S. Jhilal, E. Pasqua, C. Marchesi, R. Corradi, and M. Galletti, "Tailoring AI-Driven Reading Scaffolds to the Distinct Needs of Neurodiverse Learners," Apr. 02, 2026. doi: 10.48550/arXiv.2603.28370.